\documentclass[runningheads]{llncs}
\usepackage[T1]{fontenc}
\usepackage{graphicx}
\usepackage{amsfonts}
\usepackage{amsmath}
\usepackage{booktabs}
\usepackage{hyperref}

\begin{document}
\title{Do Recipes Have Personas? \\ Characterizing and Generating Creator Style \\ in Attributed Procedural Graphs}
%
%
\author{Lei Jiang}
\authorrunning{L. Jiang}
%
\institute{Microsoft, Redmond WA 98052, USA \\
\email{lionelchange@hotmail.com}}

\newcommand\blfootnote[1]{%
	\begingroup
	\renewcommand\thefootnote{}\footnote{#1}%
	\addtocounter{footnote}{-1}%
	\endgroup
}
\maketitle   
\blfootnote{Accepted at the 29th International Conference on Discovery Science (DS 2026). The final authenticated publication will be available online via Springer LNCS.}           
\begin{abstract}
While large language models (LLMs) possess vast zero-shot procedural knowledge, their tendency to produce homogenized logic often obscures the unique, idiosyncratic execution processes of individual human creators. In this paper, we investigate the computational discovery of procedural personas from unstructured data. To achieve this, we introduce ViralRecipesTrans, a new dataset of procedurally aligned execution flow graphs extracted from popular culinary video transcripts and explicitly mapped to specific creators. We formulate procedural stylometry as a graph learning and process discovery task, revealing a fundamental duality: while traditional lexical classifiers overfit via semantic leakage, discrete topological metrics successfully capture the rigid physical constraints of a creator's workflow. Building upon this characterization, we extend our framework into a novel generative task—predicting a creator's exact structural execution graph for unseen dishes. We expose a fundamental dichotomy in style generation between global macro-planning and local structural execution. Our results demonstrate that few-shot LLMs dominate semantic assignment but suffer from persistent macro-planning deficits, whereas our structured two-stage model achieves superior topological control via rigid Markovian priors. Together, an ensemble approach to procedural generation combines the strengths from both sides, dynamically synthesizing global semantic reasoning with localized topological footprints to automate the discovery and generation of personalized workflows. 

\end{abstract}
\section{Introduction}
Procedural text is one of the most fundamental mechanisms for transmitting complex human knowledge, making it a longstanding testbed for natural language understanding. Historically, culinary recipes were a highly common domain in Natural Language Processing (NLP) research, serving as the primary medium to evaluate event tracking, state-change modeling, and structured sequence generation~\cite{Mori2014,Bosselut2018,Marin2019,Donatelli2021}. While recipe generation tasks are usually measured by linguistic quality with metrics like BLEU~\cite{Papineni2002}, strictly text-centric recipe datasets like~\cite{Bien2020} also provided the structural scale necessary for digging into the culinary aspect with the cooking steps decomposed and compared against each other.

This capability, however, exposes a profound contradiction in the perceived problem space. Anthropologically, canonical culinary knowledge is relatively bounded, comprising roughly 20,000 established global dishes~\footnote{source: https://www.tasteatlas.com/map}. From a pure knowledge-extraction perspective, this finite space superficially appears "solved" by LLMs, where a structured representation is reliably parsed from text in tools~\cite{Mavrogiannis2024,Kopitar2025}. Yet, we observe a paradoxical, daily explosion of viral recipe videos and endless procedural variants for the exact same dishes. If the semantics of cooking is a solved problem, what drives this continuously expanding, highly engaged domain? The answer lies entirely in the human creator. While LLMs default to generating logically sound, homogenized outputs~\cite{Bansal2024}, they smooth over the very essence of human variance: the idiosyncratic structural footprints of a specific author's workflow. It is this \textit{procedural stylometry}—the latent personas governing how a task is physically approached, parallelized, and paced—that constitutes the true signal amidst the noise, a signal which can no longer be ignored if we aim to fully characterize human authorship.

In summary, culinary recipes possess several unique linguistic and structural properties that make them an ideal testbed for advancing computational stylometry given the duality of text and topology as recipes are not merely sequential text; they are fundamentally executable Directed Acyclic Graphs (DAGs). Also, the real-world execution of any single dish exhibits massive structural variation across different creators on top of the underlying causal physical and chemical constraints, making recipe data a unique environment for intra-class stylometry. 

To investigate these fingerprints on both linguistics and structure, we introduce the \textit{ViralRecipesTrans} (VRT) dataset\footnote{Dataset and code in this paper: \url{https://github.com/lionelc/RecipePersonas}} in Section~\ref{dataset}, a novel corpus (along with procedurally aligned DAGs extracted) from the transcripts of viral culinary video creators. After a rigorous empirical analysis of procedural stylometry in Section~\ref{recipepersonas}, we consider the interesting generative task \textit{"creating a viral recipe with the given dish name and ingredients"}, where the dichotomy between global semantic macro-planning and local procedural execution is conceptualized in Section~\ref{gen-section}. Specifically, the following core research questions are  addressed:

\textbf{RQ1 (Lexical vs Topology in Procedural Personas)}: How are author-specific stylometric signatures distributed across textual expression and structural topology, and to what extent can analyzing this duality overcome the overfitting of purely lexical classification?

\textbf{RQ2 (Persona-Conditioned Graph Generation)}: How effectively can a generative framework predict the complete procedural graph of a recipe—including task ordering and structural ellipsis—conditioned solely on the target dish, base ingredients, and the creator’s characterized persona?

\textbf{RQ3 (Global vs. Local Structural Modeling)}: How do the inherent trade-offs between global semantic reasoning and local Markovian constraints manifest across different generative architectures, and can an adaptive ensemble framework successfully synthesize these complementary forces to maximize operational fidelity?

\section{Related Work}

\subsection{Computational Stylometry and Persona Modeling}
Historically, computational stylometry has relied on surface-level lexical features, such as N-grams~\cite{Brown1992} and TF-IDF~\cite{Jones1972}, to determine authorship~\cite{Stamatatos2009}. Because these discriminative methods severely overfit to vocabulary in constrained domains, recent work has explored structural signatures, utilizing neural models on linguistic dependency trees to withstand cross-topic variance~\cite{Jafariakinabad2019}. Concurrently, persona modeling has evolved into a generative task. While models trained on conversational datasets frame personas as explicit semantic facts~\cite{Gu2021}, text style transfer frameworks largely treat persona as a surface-level lexical overlay, modifying tone or vocabulary rather than underlying logic~\cite{Jin2022}. Even within the culinary domain, existing personalization efforts focus on modeling consumer semantic preferences, such as generating recipes with user-favored ingredients~\cite{Majumder2019}. Our work departs from these paradigms. Because culinary procedures artificially compress semantic variance through physical logic, we introduce procedural stylometry. Rather than classifying explicit traits or generating lexical substitutions, we investigate the latent workflow signatures of creators, formulating persona modeling as a challenge of structural style transfer over execution graphs.

\subsection{Controlled Generation and Style Transfer}
The advent of Retrieval-Augmented Generation (RAG)~\cite{Lewis2020} has increasingly positioned LLM as parametric knowledge bases~\cite{Petroni2019}, where models generate zero-shot contextual frameworks prior to reasoning~\cite{Yu2023}. While this paradigm excels at procedural extraction, enforcing specific topological structures during generation remains a persistent challenge. Existing frameworks~\cite{Shen2017,Rao2018,Lample2019} for controllable generation typically employ neural logic constraints~\cite{Lu2021} to ensure the output adheres to valid physical rules. However, these methods treat structural control purely as a mechanism for objective accuracy—ensuring a generated recipe is executable. In contrast, our work introduces subjective structural style transfer. Within NLP, structural style transfer has recently emerged via benchmarks like StylePTB~\cite{Lyu2021} and methodologies such as optimal transport~\cite{Nouri2022}, requiring models to manipulate compositional ordering while preserving semantic intent~\cite{Akyurek2023}. Yet, adapting LLMs for this task is difficult; without explicit architectural intervention, zero-shot LLMs frequently collapse into structural copying, failing to induce the target style~\cite{Lai2024}. Our framework unites these two isolated research tracks.

\section{The ViralRecipesTrans (VRT) Dataset: Construction and Statistics}
\label{dataset}
To empirically investigate the phenomenon of procedural stylometry, the VRT dataset comprises video transcripts mapped to execution DAGs across culinary creators. While there exist real-world limitations on the total number of cooking influencers with qualified video formats (with valid transcripts and English-language narratives), we argue that it contains adequate persona information as the results in Sections~\ref{recipepersonas} and~\ref{ablation} suggest. This dataset also presents a contrast to the standardized procedural text found in traditional recipe corpora. 

\subsection{Account Discovery and Filtering}
To construct the dataset, we implemented an automated discovery pipeline using the YouTube Data API. Starting from layered seed queries (e.g., general cooking $\to$ food-type $\to$ cuisine-specific), the pipeline identified candidate channels and applied a rigorous multi-stage filter to ensure data density and procedural diversity:

\begin{itemize}
	\item \textbf{Volume and Reach}: Channels must possess a minimum of 10,000 subscribers and at least 50 uploaded videos.
	\item \textbf{Format Constraints}: At least 30\% of recent uploads must be long-form content (greater than 3 minutes in duration).
	\item \textbf{Repertoire Diversity}: An LLM-based dish detection pass on video titles was required to identify $\geq 30$ unique dishes per account, ensuring recipe diversity.
	\item \textbf{Transcript Quality}: Three randomly sampled videos per channel were required to yield English transcripts with sufficient word counts and successfully compile into valid procedural graphs.
\end{itemize}

The combined dataset spans 97 unique accounts (with 3 overlapping between long and short formats), ranging from niche home cooks (14K subscribers) to major culinary channels (7.7M subscribers), totaling more than 5,000 videos. 

\subsection{Transcript Collection}
For each qualifying account, we fetched English transcripts for up to 50 recipe videos per format, ranked by total view count. Transcripts were obtained via the youtube-transcript-api. To ensure pipeline stability and prevent connection hangs, requests were throttled with a 20–40 second delay and a 90-second signal-based timeout.

Each transcript was quality-checked by building a flow action graph via GPT-5.4 and verifying $\ge$2 cooking actions and $\ge$2 distinct ingredients. Low-quality transcripts were discarded automatically.

\subsection{Descriptive Statistics}
The VRT dataset establishes a robust environment for studying procedural style transfer. The global action vocabulary extracted from the transcripts comprises 570 unique cooking verbs and exhibits a heavy long-tailed distribution. The top five most frequent actions (add, mix, cook, stir, cut) account for 37\% of all action tokens, while 47\% of the per-account actions are hapax legomena (verbs used only once by that specific creator).

Detailed corpus and per-recipe statistics are provided in Table~\ref{dataset-t1}.

\begin{table*}
	\centering
	\caption{VRT per-recipe textual and topological statistics}
	\label{dataset-t1} 
	\begin{tabular}{llll}
		\hline
		\textbf{Metric}           & \textbf{Long-form}  & \textbf{Short-form}  & \textbf{Overall} \\
		\hline
		Cooking steps (mean $\pm$ std)      &  14.8 $\pm$ 10.6   &    9.3 $\pm$ 4.8   &  12.2 $\pm$ 8.9 \\
		Unique Ingredients (mean $\pm$ std)    &     13.2 $\pm$ 8.0 &    10.0 $\pm$ 5.2    & 11.7 $\pm$ 7.0                  \\
		Transcript length (words, mean)      &   1800  & 176    & 1028                    \\
		Transcript length (words, median)  &    1310 &  153      & 299                      \\
		Global action vocabulary   &    548  &  499    &  675    \\
		Action vocab per account (mean) & 106  & 89  & 97\\
		Hapax ratio (per-account mean)  & 40\%  &  46\%  & 43\% \\
		Auto-generated captions & 65\% & 96\%  &  80\%  \\
		\hline
	\end{tabular}
\end{table*}

\section{Detecting and Characterizing Recipe Personas}
\label{recipepersonas}
Before evaluating fine-grained authorship, we first establish that procedural topology captures the macro-level constraints of the medium. If structural graphs genuinely reflect operational behavior, they should natively differentiate between foundational procedural archetypes: traditional written text (from food.com~\cite{Majumder2019} and RecipeNLG~\cite{Bien2020}), narrative long-form and compressed short-form videos.

\subsection{Archetype-Level Structural Classification}
To evaluate macro-level procedural topologies, we formally define the extraction of structural archetypes. Let a recipe be represented as an attributed DAG $\mathcal{G} = (\mathcal{V}, \mathcal{E}, \mathcal{X})$. We define a purely topological projection function $\phi: \mathcal{G} \to \mathbb{R}^d$, which strips the semantic matrix $\mathcal{X}$ and extracts $d$ discrete graph metrics (e.g., depth-to-width ratio, branch rate, maximum layer depth). We then formulate a classification function $f_\theta: \mathbb{R}^d \to \mathcal{A}$, mapping the topology to a discrete set of medium-constrained archetypes $\mathcal{A} = \{a_{\text{text}}, a_{\text{long}}, a_{\text{short}}\}$. Using a \textit{random forest classifier}, we evaluate whether $\phi(\mathcal{G})$ contains sufficient signal to isolate these archetypes. As detailed in Table \ref{tab:archetype_results}, the model successfully separates the topological spaces, achieving an overall Macro F1 score of 0.783.

\begin{table}[t]
	\centering
	\caption{Classification performance for the 3-class procedural archetype task. The perfect isolation of the written archetype confirms that cinematographic workflows possess a fundamentally distinct structural topology.}
	\label{tab:archetype_results}
	\begin{tabular}{lccc}
		\toprule
		\textbf{Archetype Class ($\mathcal{A}$)} & \textbf{Precision} & \textbf{Recall} & \textbf{F1-Score} \\
		\midrule
		Written Text ($a_{\text{text}}$) & 0.999 & 0.999 & 0.999 \\
		Viral Long-Form ($a_{\text{long}}$) & 0.702 & 0.613 & 0.654 \\
		Viral Short-Form ($a_{\text{short}}$) & 0.656 & 0.740 & 0.696 \\
		\midrule
		\textbf{Overall Metric} & \multicolumn{3}{c}{\textbf{Score}} \\
		\midrule
		Accuracy & \multicolumn{3}{c}{0.784} \\
		Macro F1 & \multicolumn{3}{c}{0.783} \\
		\midrule
		\multicolumn{4}{l}{\textbf{Top Discriminative Features ($\phi$)}} \\
		\midrule
		\multicolumn{4}{l}{1. Depth-to-Width Ratio (0.152)} \\
		\multicolumn{4}{l}{2. Average Out-Degree (0.129)} \\
		\multicolumn{4}{l}{3. Ingredient Branch Rate (0.128)} \\
		\bottomrule
	\end{tabular}
\end{table}

Unsurprisingly, the classifier perfectly isolates the traditional written archetype ($a_{\text{text}}$) from both video formats. Feature importance analysis reveals that this boundary is heavily governed by the graph's depth and branching behavior, exposing a distinct cinematographic decomposition pattern. Written recipes exhibit shallow, highly parallelized graphs. Conversely, video-derived workflows are structurally elongated; creators deliberately fracture standard procedures into sequential, deep micro-actions to sustain visual narration, resulting in a massive increase in ingredient reuse (branch rate).

Furthermore, the topological projection captures the inherent constraints of short-form media. While short-form graphs ($a_{\text{short}}$) share the deep, sequential framework of their long-form counterparts, they exhibit a heavy compression archetype. This compression alters not just the graph size, but the operational behavior itself, as short-form workflows drastically prune complex preparatory techniques in favor of linear, "dump-and-bake" assembly sequences. This robust structural separation confirms that procedural DAGs successfully capture the overarching physical constraints of a medium. 

\subsection{The Illusion of Lexical Stylometry}
Before isolating a creator's structural workflow, we first establish the baseline capabilities of traditional linguistic stylometry on procedural transcripts. Using a 50-class authorship attribution (classification) task, we evaluated a purely lexical model (character $N$-gram TF-IDF~\cite{Stamatatos2009}) and a handcrafted stylometric model incorporating psycholinguistic features. Linguistic stylometry achieves remarkably high numerical performance. The combined model (with stratified 5-fold cross-validation) achieves a Top-1 Accuracy of 0.9006, while the purely lexical TF-IDF model peaks at 0.9357, identifying several creators with perfect precision. 

While our analysis uses raw transcripts rather than text distortion~\cite{Stamatatos2017},
which isolates personal writing style for better control, high discriminative accuracy still does not establish that the model has captured procedural style. Lexical features conflate \emph{repertoire} (what a creator cooks) with \emph{procedure} (how they cook it). This distinction is decisive for our generative objective in Section~\ref{gen-section}: there, the dish and its ingredients are supplied as input, so any signal derived from niche-specific vocabulary is unavailable by construction. Only structural signal transfers to unseen dishes.

\subsection{The Scalability of Procedural Personas}
Before evaluating the scalability of individual personas, we must ensure our classifier is learning genuine operational behavior rather than memorizing domain-specific vocabulary. As demonstrated previously, lexical baselines artificially inflate performance by exploiting "lexical leakage"—memorizing the specific ingredients or linguistic quirks associated with a creator's preferred cuisine. To rigorously eliminate this leakage, we strip our graphical representation down to a strict, 19-dimensional pure topological feature space, discarding all semantic node labels, action bigram transitions, and specific verb categories. Focusing on structural persona, we formulate a scalable authorship attribution task, varying the class space from $K=2$ up to $K=50$ creators, utilizing 5 randomized sampling trials per $K$ to prevent dish-overfitting. 

\begin{figure*}[t]
	\includegraphics[width=1.0\linewidth]{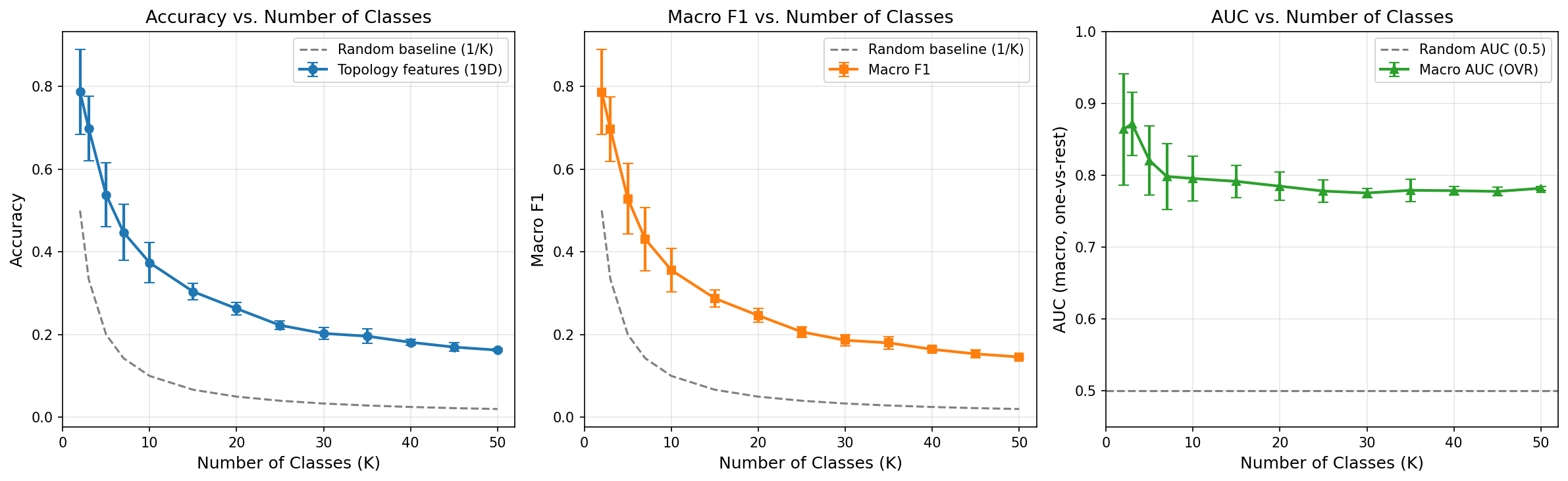}
	\caption {Scalability results as the number of classes increases}
	\label{scalability}
\end{figure*}

The results, visualized in Figure~\ref{scalability}, reveal a fundamental divergence between rigid boundary classification and latent signal strength.

\subsubsection{Sensitivity to Feature Dimensionality}
As the classification space scales, absolute Top-1 Accuracy and Macro F1 naturally decay. However, this downward trajectory is an artifact of the classification mechanism itself, not a degradation of the underlying persona signal. Cooking is governed by strict physical logic; as we crowd the feature space with 50 different workflows, the shared baseline mechanics (e.g., all chefs must apply heat after preparation) cause the rigid discriminative hyperplanes to overlap and shatter. Yet, when evaluated against a random baseline ($1/K$), the model's relative predictive power actually accelerates as complexity increases. The growing delta between the empirical accuracy curve and the random baseline confirms that the model is isolating a genuine, persistent structural footprint rather than exploiting low-resource noise.

\subsubsection{The Probabilistic Nature of Persona}
The true nature of this structural footprint is exposed by the Area Under the ROC Curve (AUC). As shown in the rightmost panel of Figure~\ref{scalability}, while hard classification boundaries fail, the AUC remains remarkably flat and stable—hovering near 0.78 even at $K=50$. This stability proves that the procedural persona operates as a probabilistic ranking signal rather than a strict categorical boundary. Even when the model cannot draw a perfect geometric boundary around a creator's specific style, it retains a deep enough mathematical understanding of their workflow to consistently rank them near the very top of a massive candidate distribution.

This empirical divergence formally dictates our subsequent methodology. Because human procedural variance exists as a robust distributional preference rather than a static classification boundary, discriminative classifiers ultimately hit a representational ceiling. To fully characterize, capture, and replicate the idiosyncratic routing of a specific creator, we also need to move forward onto generative modeling.

\section{Modeling Procedural Variance for Persona Generation}
\label{gen-section}
Having established that procedural personas exist as latent structural variances rather than strict discriminative boundaries, we pivot to the generative challenge. To prove that we can successfully isolate and model a creator's workflow, we must be able to autonomously reconstruct how that specific creator would physically approach an unseen recipe.

\subsection{Task Formulation}
We formally define the persona-conditioned generation task as a graph prediction problem. Let a canonical recipe be represented by its semantic dish name $D$ and a set of raw ingredients $I$. Given a target creator persona $C$, our objective is to generate an execution graph $\hat{\mathcal{G}}_C = (\mathcal{V}, \mathcal{E})$ that replicates the exact physical workflow the creator would employ. Unlike standard conditional text generation (e.g., "rewrite this recipe in the style of $C$"), this task does not require predicting the creator's spoken dialogue or vocabulary. Instead, the model must predict the sequence of action states and the precise routing of ingredients into those states. A successful generation $\hat{\mathcal{G}}_C$ must reflect the creator's distinct topological signature—matching their typical depth, branching patterns, and ingredient usage—while maintaining the physical constraints of the dish $D$.

\begin{figure}[t]
	\includegraphics[width=0.9\linewidth]{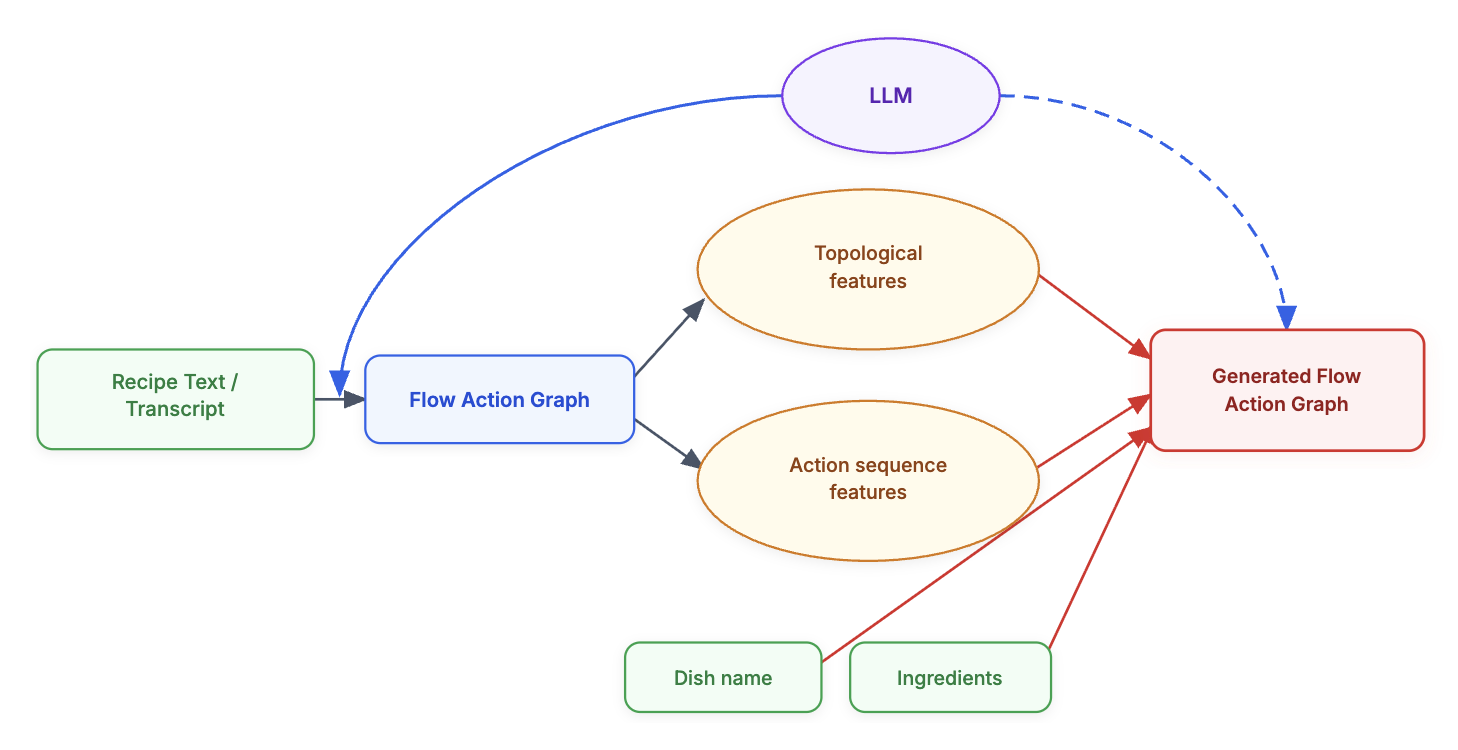}
	\caption {The workflow of the persona-controlled recipe generation task}
	\label{gen-task}
\end{figure}

Figure~\ref{gen-task} shows the workflow of the generative task along with LLM's role: in the preprocessing of the extraction of flow action graphs, LLM has proven to be robust. Meanwhile, with all the features deterministically constructed, LLM could help with world knowledge for an unknown dish. 

\subsection{Evaluation Paradigm: Physics over Text}
Traditional text-generation tasks in the culinary domain heavily rely on linguistic overlap metrics like BLEU~\cite{Papineni2002}  or ROUGE~\cite{Lin2004}: for transcript-style text, its representational power is not as strong as shown in Section~\ref{recipepersonas}. To evaluate whether a generated graph $\hat{\mathcal{G}}$ accurately captures the creator's persona, we abandon text-overlap metrics in favor of topological alignments against the creator's LLM-extracted execution graph $\mathcal{G}$ as a given target (which has a high reliability). We introduce two primary metrics inspired by Dice similarity coefficient:

\textbf{Ingredient F1 (IF1)}: Creators frequently alter standard recipes by omitting steps or adding signature components. To isolate the model's semantic assignment capabilities from its topological planning, we measure IF1 strictly on the subset of steps where the predicted physical action aligns with the ground truth. Let $S_{align}$ be the set of steps where $\hat{a}_t = a_t$. We define IF1 as the average harmonic mean of precision and recall between the predicted ingredient subset $\hat{W}_t$ and the actual ingredient subset $W_t$ for all aligned steps:

\[
\text{IF1} = \frac{1}{|S_{align}|} \sum_{t \in S_{align}} \frac{2 \cdot |\hat{W}_t
	\cap W_t|}{|\hat{W}_t| + |W_t|}
\]

\textbf{Normalized Edge F1 (nEF1)}: This is our primary metric for operational fidelity. It evaluates whether the model routed the correct ingredient into the correct physical state. We extract the set of directed edges $E$, where each edge is a tuple $(a, w)$ representing an ingredient $w$ being subjected to action $a$. To ensure we are measuring physics rather than vocabulary (lexical leakage), we apply a normalization function $norm(a)$ that maps synonymous verbs (e.g., "fry", "sauté", "sear") to a canonical physical state ("cook"). The Normalized Edge F1 is calculated as the overlap of these state-ingredient tuples between the predicted and actual graphs:

\[
\text{nEF1} = \frac{2 \cdot |\hat{E}_{norm} \cap E_{norm}|}{|\hat{E}_{norm}| + |E_{norm}|}
\]

By requiring the model to match the exact $(a, w)$ tuples, nEF1 heavily penalizes structural hallucinations. A high nEF1 guarantees that the model has not just mimicked the creator's tone, but successfully replicated their distinct operational footprint, making it a very relevant metric in culinary rigor. 

\textbf{Macro-Complexity and Sequence Alignments}: In addition to ingredient and edge mapping, we evaluate the macroscopic length prediction using Step Error, defined as the absolute normalized difference between the predicted and actual number of procedural steps: $\text{Step Error} = \frac{|\hat{N} - N|}{N}$. Also, to measure pure action selection and ordering independent of ingredients, we track the \textit{Action Jaccard Index} and the \textit{Longest Common Subsequence} Ratio between the predicted and actual action sequences, where empirical analysis shows they correlate heavily with nEF1. 

\subsection{The Spectrum of LLM-Assisted Generative Baselines}
To establish a robust methodology for procedural graph generation, we evaluate a series of generative paradigms, moving from macroscopic semantic prompting to microscopic topological unrolling. Naturally, when generating an unseen dish, the \textit{world knowledge} needed is where LLM can help. 

\textbf{Zero-Shot Generation}: The most naive baseline treats persona replication as a standard text-based style-transfer task. Given the dish $D$ and ingredients $I$, an LLM is prompted zero-shot to output the execution graph in the target creator's style. 

\textbf{Few-Shot In-Context Learning}: a few-shot paradigm, providing the LLM with a retrieved subset of the creator's historical execution graphs as in-context exemplars, along with a retrieved formal recipe in text.

\textbf{Microscopic Unrolling (Step-by-Step Prediction)}: To mitigate macroscopic planning failures, we decompose the graph generation into a strictly step-by-step autoregressive process. That is, the system is tasked only with predicting the next sequential action $a_t$ and its associated ingredient subset $w_t$, strictly conditioned on the executed state history. Within this step-by-step framework, we isolate two distinct predictive engines: \textbf{Autoregressive LLM}: This engine relies on semantic few-shot prompting to predict the next logical step; and correspondingly, \textbf{N-Gram Guided Topology} abandons semantic reasoning for the action-prediction phase. It utilizes the creator's pure statistical action transition matrix to predict the next physical state $P(a_t | a_{t-1})$. 

Also, as different generators show a variance on dishes, an ensemble method is also implemented with a~\textbf{selector} trained over a subset of features and based on other models' performance in training. 

\subsection{Structured Two-Stage Generation: Disentangling Skeleton and Semantics}
While the LLM-driven approaches provides strong semantic reasoning, it relies entirely on black-box, token-intensive API calls at every step. We also introduce a fully transparent, LLM-free (at inference) generative algorithm: Structured Two-Stage (S2S) Generation~\cite{Fan2018,Yao2019}.

This approach mathematically disentangles the generation of the structural topology (the skeleton) from the ingredient assignment (the semantics), allowing us to apply learned creator-specific statistical priors at both levels.

\subsubsection{Stage 1: Topological Skeleton Generation (Action Planning)}
The first stage generates the sequence of action states $A = (a_1, a_2, \dots, a_N)$. Rather than relying on an LLM to blindly guess the length of the workflow, we deterministically predict the macro-complexity $N$. The step count is calculated via a learned heuristic that fuses the creator's historical mean step count, their ratio of viral-steps to formal-steps, and their steps-per-ingredient density.

With $N$ established, we employ Beam Search to decode the optimal action sequence. At each step $t$, a candidate action $a_t$ is evaluated using a composite logarithmic objective function that dynamically fuses the creator's physical habits with the semantic requirements of the dish:
\[
\text{Score}(a_t) = \alpha \log P_{trans} + \beta \log P_{sem} +
\]
$$ \gamma \log P_{pos} + \delta \log P_{uni} - \lambda(A)$$

Where $P_{trans}(a_t | a_{t-1})$: The creator's learned bigram transition prior (the physics); $P_{sem}(a_t | \mathcal{F}_t, I)$: The semantic alignment prior, derived from mapping the formal recipe's actions $\mathcal{F}$ to the creator's viral vocabulary via historical ingredient overlap;$P_{pos}(a_t | \phi_t)$: A positional prior measuring the creator's likelihood to use $a_t$ in the current phase $\phi_t \in \{\text{early, mid, late}\}$; and $\lambda(A)$: A structural penalty applied to highly repetitive, stagnant action loops.

\subsubsection{Stage 2: Semantic Flesh (Ingredient Assignment)}
Once the skeleton $A$ is generated, the algorithm deterministically routes the available raw ingredients $I$ to the predicted nodes. Instead of relying on LLM semantic reasoning, the assignment is driven by a learned bipartite mapping. An ingredient $w$ is routed to action $a_t$ if it maximizes the creator's historical ingredient-action prior $P(a_t | w)$ and exhibits high lexical overlap with the aligned formal step. To preserve the creator's topological depth-to-width ratio, we mathematically bound the maximum fan-in (ingredients per step).

\subsection{Performance Comparisons}

\begin{table*}[b!]
	\centering
	\caption{Performance comparison over the stylish recipe generation task}
	\label{genperf}
	\begin{tabular}{l c c c c c c}
		\toprule
		& \multicolumn{3}{c}{\textbf{long videos}} & \multicolumn{3}{c}{\textbf{short videos}} \\
		\cmidrule(lr){2-7}
		\textbf{Item} & \textbf{IF1} & \textbf{nEF1} & \textbf{StepErr} & \textbf{IF1} & \textbf{nEF1} & \textbf{StepErr} \\
		\midrule
		Zero-shot LLM & 0.223 & 0.338 & 0.481 & 0.261 & 0.341 & 0.458 \\
		Few-shot LLM & \textbf{0.344} & 0.474 & 0.373 & \textbf{0.385} & 0.478 & 0.351 \\
		Autoregressive & 0.318 & 0.473 & 0.446 & 0.353 & 0.446 & 0.410 \\
		N-gram sequence & 0.303 & 0.452 & 0.549 & 0.363 & 0.446 & 0.513 \\
		Two-stage & 0.294 & 0.456 & \textbf{0.350} & 0.348 & 0.455 & \textbf{0.332} \\
		Ensemble with selector & 0.328 & \textbf{0.485} & 0.357 & 0.378 & \textbf{0.479} & 0.352 \\
		\bottomrule
	\end{tabular}
\end{table*}

An analysis of the generative baselines in Table~\ref{genperf} reveals a stark contrast in where different architectures excel. When evaluating semantic assignment, the few-shot LLM significantly outperforms the LLM-free baselines, securing the highest Ingredient F1 (IF1) on both long (0.344) and short (0.385) video formats compared to S2S (0.294 and 0.348, respectively). Conversely, when evaluating macro-structural planning, S2S is the definitive leader. By strictly enforcing topological sequences, it achieves the lowest Step Error across both long (0.350) and short (0.332) videos, cleanly beating the few-shot baseline (0.373 and 0.351). Despite these opposing strengths, both models achieve highly competitive overall operational fidelity; the few-shot model reaches an nEF1 of 0.474 (long) and 0.478 (short), while S2S reaches 0.456 (long) and 0.455 (short). Notably, intermediate microscopic approaches fail to dominate in any metric, because forcing generation into isolated step predictions without planning actually fractures the LLM’s overarching contextual logic. Substituting a different creator's priors while holding the dish and ingredients fixed costs S2S 0.06 nEF1 on both formats ($p<10^{-54}$), confirming that this fidelity derives from persona conditioning rather than generic culinary structure.

These metrics demonstrate that both models reach comparable operational fidelity through entirely divergent, yet complementary, mechanisms. The few-shot baseline leverages its massive parametric memory to dominate semantic assignment, seamlessly routing complex ingredients into the graph. However, it remains constrained by the inherent macro-planning deficits of autoregressive generation, resulting in elevated Step Errors. In contrast, S2S model relies on explicit Markovian priors~\cite{Perez2024} to perfectly control the sequence topology, securing rigid structural adherence while naturally sacrificing the zero-shot semantic agility of an LLM. This empirical trade-off proves that no singular paradigm can simultaneously maximize semantic nuance and structural rigidity. Because capturing a complete procedural persona requires both global semantic awareness and local topological control, the complementary strengths directly motivate an ensemble framework, which utilizes a learned selector to dynamically route generation to a certain model.

\begin{table*}[b!]
	\centering
	\caption{Ablation study on LLM backbone replacement: comparison of generative performance across proprietary frontier and open-weight models.}
	\label{tab:llm_backbone_ablation}
	\begin{tabular}{l c c c c c c}
		\toprule
		& \multicolumn{3}{c}{\textbf{Long Videos}} & \multicolumn{3}{c}{\textbf{Short Videos}} \\
		\cmidrule(lr){2-4} \cmidrule(lr){5-7}
		\textbf{Model} & \textbf{IF1} & \textbf{nEF1} & \textbf{StepErr} $\downarrow$ & \textbf{IF1} & \textbf{nEF1} & \textbf{StepErr} $\downarrow$ \\
		\midrule
		GPT-5.4       & \textbf{0.396} & \textbf{0.556} & 0.384 & 0.413 & \textbf{0.532} & 0.317 \\
		GPT-4.1-mini  & 0.393 & 0.504 & \textbf{0.376} & \textbf{0.481} & 0.446 & 0.269 \\
		Qwen3.6-27B   & 0.369 & 0.510 & 0.409 & 0.410 & 0.476 & \textbf{0.248} \\
		\bottomrule
	\end{tabular}
\end{table*}

\section{Ablation Study}
\label{ablation}
\textbf{LLM backbone selection}: in Table~\ref{tab:llm_backbone_ablation}, substituting the core LLM engine across both proprietary and open-weight architectures proves that raw parameter scaling doesn't resolve the macro-planning deficit. A larger model is not necessarily better at adhering to rigid procedural topologies. 

\textbf{Context window saturation in few-shot prompting}: To evaluate the few-shot baseline's reliance on in-context style exemplars, the number of shots ($k$) is scaled in order to isolate how much of the creator's procedural footprint is absorbed dynamically versus relying on foundational knowledge. While increasing exemplars yields steady improvements in semantic assignment (peak performance at $k=9$), injecting too many complex procedural graphs causes context dilution (starting from $k=11$), overwhelming LLM's ability to effectively attend to key stylistic signals.

\section{Conclusion and Future Work}
In this work, we challenge the prevailing NLP paradigm for procedural style transfer by demonstrating that traditional linguistic stylometry is inadequate in persona-controlled recipe generation: considering procedural style as a topological footprint, encoding  a creator's persona in the physical shape of their execution graph rather than their vocabulary, we introduced the ViralRecipesTrans dataset along with metrics to measure operational fidelity, alongside a novel structured two-stage generation algorithm. Between this model’s macro-planning superiority and LLM’s micro-semantic agility from few-shot scenarios, the trade-off is explored with insight on an ensemble. 

Future work will bridge this gap by routing toward true neuro-symbolic fusion with interoperability as well as a post-hoc calibrator, which empirically validates physical constraints and repairs potential violations. Further, while this framework was built in the culinary domain, the concept of "topological stylometry" is broadly applicable in other procedural domains for its graph-theoretic extraction, moving toward generalized, persona-aware instructional intelligence.

%
%
%

\begin{thebibliography}{8}
\bibitem{Akyurek2023}
Ekin Akyürek, Afra Amini, Jacob Andreas, and Yoon Kim: Compositional Generalisation with Structured Reordering and Fertility Layers. on Proceedings of the Conference of the European Chapter of the Association for Computational Linguistics (EACL), pp. 2157--2171 (2023).

\bibitem{Bansal2024}
Hritik Bansal, Ran Levy, and Yonatan Bisk: Order-based pre-training strategies for procedural text understanding. on Proceedings of the Conference of the North American Chapter of the Association for Computational Linguistics: Human Language Technologies (NAACL), pp. 645--652 (2024).
	
\bibitem{Bien2020}
Micha{\l} Bien, Micha{\l} Gilski, Maciej Maciejewska, Wojciech Taisner, Dawid Wisniewski, and Agnieszka Lawrynowicz: RecipeNLG: A Cooking Recipes Dataset for Semi-Structured Text Generation.
on Proceedings of the International Conference on Natural Language Generation, pp. 22--28 (2020).

\bibitem{Bosselut2018}
Antoine Bosselut, Omer Levy, Ari Holtzman, Corin Ennis, Dieter Fox, and Yejin Choi: Simulating Action Dynamics with Neural Process Networks. on Proceedings of the International Conference on Learning Representations (2018).

\bibitem{Brown1992}
Peter F. Brown, Vincent J. Della Pietra, Peter V. deSouza, Jenifer C. Lai, and Robert L. Mercer: Class-Based n-gram Models of Natural Language. on Computational Linguistics, 18(4), pp. 467--480 (1992).

\bibitem{Donatelli2021}
Lucia Donatelli, Theresa Schmidt, Debanjali Biswas, Arne K\"{o}hn, Fangzhou Zhai, and Alexander Koller: Aligning Actions Across Recipe Graphs. on Proceedings of the Conference on Empirical Methods in Natural Language Processing (EMNLP), pp. 6930--6942 (2021).

\bibitem{Fan2018}
Angela Fan, Mike Lewis, and Yann Dauphin: Hierarchical Neural Story Generation. on Proceedings of the Annual Meeting of the Association for Computational Linguistics (ACL), pp. 889--898 (2018).

\bibitem{Gu2021}
Jia-Chen Gu, Zhen-Hua Ling, Quan Liu, Zhigang Chen, and Xiaodan Zhu: Detecting speaker personas from conversational texts. on Proceedings of the Conference on Empirical Methods in Natural Language Processing, pp. 1126--1136 (2021).

\bibitem{Jafariakinabad2019}
Faezeh Jafariakinabad, Sansiri Tarnpradab, and Kien A. Hua: Syntactic recurrent neural network for authorship attribution. on Proceedings of the IEEE International Conference on Semantic Computing (ICSC), pp. 251--258 (2019).

\bibitem{Jin2022}
Di Jin, Zhijing Jin, Zhiting Hu, Vlas Vovk, and Liwei Wang: Deep learning for text style transfer: A survey. on Computational Linguistics, 48(1):155--205 (2022).

\bibitem{Jones1972}
Karen Spärck Jones: A Statistical Interpretation of Term Specificity and Its Application in Retrieval. on Journal of Documentation, 28(1), pp. 11--21 (1972).

\bibitem{Kopitar2025}
Leon Kopitar, Leon Bedra\v{c}, Larissa J. Strath, Jiang Bian, and Gregor Stiglic: Improving Personalized Meal Planning with Large Language Models: Identifying and Decomposing Compound Ingredients. Nutrients, 17(9), 1492 (2025).

\bibitem{Lai2024}
Wen Lai, Viktor Hangya, and Alexander Fraser: Style-Specific Neurons for Steering LLMs in Text Style Transfer. on Proceedings of the Conference on Empirical Methods in Natural Language Processing (EMNLP), pp. 5410--5425 (2024).

\bibitem{Lample2019}
Guillaume Lample, Sandeep Subramanian, Eric Smith, Ludovic Denoyer, Marc'Aurelio Ranzato, and Y-Lan Boureau: Multiple-Attribute Text Rewriting. on Proceedings of the International Conference on Learning Representations (ICLR) (2019).

\bibitem{Lewis2020}
Patrick Lewis, Ethan Perez, Aleksandara Piktus, Fabio Petroni, Vladimir Karpukhin, Naman Goyal, Heinrich Küttler, Mike Lewis, Wen-tau Yih, Tim Rocktäschel, Sebastian Riedel, and Douwe Kiela: Retrieval-Augmented Generation for Knowledge-Intensive NLP Tasks. on Advances in Neural Information Processing Systems, 33, pp. 9459--9474 (2020).

\bibitem{Lin2004}
Chin-Yew Lin: ROUGE: A Package for Automatic Evaluation of Summaries. on Text Summarization Branches Out, pp. 74--81 (2004).

\bibitem{Lu2021}
Ximing Lu, Peter West, Rowan Zellers, Ronan Le Bras, Chandra Bhagavatula, and Yejin Choi: NeuroLogic Decoding: (Un)directed Generation with Neural Logic Constraints. on Proceedings of the Conference of the North American Chapter of the Association for Computational Linguistics: Human Language Technologies, pp. 4287--4299 (2021).

\bibitem{Lyu2021}
Yiwei Lyu, Paul Pu Liang, Hai Pham, Eduard Hovy, Barnabás Póczos, Ruslan Salakhutdinov, and Louis-Philippe Morency: StylePTB: A Compositional Benchmark for Fine-grained Controllable Text Style Transfer. on Proceedings of the Conference of the North American Chapter of the Association for Computational Linguistics: Human Language Technologies (NAACL), pp. 2116--2138 (2021).

\bibitem{Majumder2019}
Bodhisattwa Prasad Majumder, Shuyang Li, Jianmo Ni, and Julian McAuley: Generating personalized recipes from historical user preferences. on Proceedings of the Conference on Empirical Methods in Natural Language Processing (EMNLP), pp. 5976--5982 (2019).

\bibitem{Marin2019}
Javier Marin, Aritro Biswas, Ferda Ofli, Nicholas Hynes, Amaia Salvador, Yusuf Aytar, Ingmar Weber, and Antonio Torralba: Recipe1M+: A Dataset for Learning Cross-Modal Embeddings for Cooking Recipes and Food Images. on IEEE Transactions on Pattern Analysis and Machine Intelligence, 43(1):187--203 (2019).

\bibitem{Mavrogiannis2024}
Angelos Mavrogiannis, Christoforos Mavrogiannis, and Yiannis Aloimonos: Cook2LTL: Translating Cooking Recipes to LTL Formulae using Large Language Models. on Proceedings of IEEE International Conference on Robotics and Automation (ICRA), 17679--17686 (2024).

\bibitem{Mori2014}
Shinsuke Mori, Hirokuni Maeta, Tetsuro Sasada, Koichiro Yoshino, Yoko Yamakata, and Tetsuji Itoh: Flow Graph Corpus from Recipe Texts. on Proceedings of the Ninth International Conference on Language Resources and Evaluation (LREC), 2370--2377 (2014).

\bibitem{Nouri2022}
Nasim Nouri: Text Style Transfer via Optimal Transport. on Proceedings of the Conference of the North American Chapter of the Association for Computational Linguistics: Human Language Technologies (NAACL), pp. 2532--2541 (2022).

\bibitem{Papineni2002}
Kishore Papineni, Salim Roukos, Todd Ward, and Wei-Jing Zhu: Bleu: a Method for Automatic Evaluation of Machine Translation. on Proceedings of the Annual Meeting of the Association for Computational Linguistics (ACL), pp. 311--318 (2002).

\bibitem{Perez2024}
Guillaume Perez et al.: Markov Constraint as Large Language Model Surrogate. on Proceedings of the International Joint Conference on Artificial Intelligence (IJCAI), pp. 1844--1852 (2024).

\bibitem{Petroni2019}
Fabio Petroni, Tim Rocktäschel, Patrick Lewis, Anton Bakhtin, Yuxiang Wu, Alexander H. Miller, and Sebastian Riedel: Language Models as Knowledge Bases? on Proceedings of the Conference on Empirical Methods in Natural Language Processing and the 9th International Joint Conference on Natural Language Processing (IJCNLP), pp. 2463--2473 (2019).

\bibitem{Rao2018}
Sudha Rao and Joel Tetreault: Dear Sir or Madam, May I Introduce the GYAFC Dataset: Corpus, Benchmarks and Metrics for Formality Style Transfer. on Proceedings of the Conference of North American Chapter of the Association for Computational Linguistics: Human Language Technologies (NAACL), pp. 129--140 (2018).

\bibitem{Shen2017}
Tianxiao Shen, Tao Lei, Regina Barzilay, and Tommi Jaakkola: Style Transfer from Non-Parallel Text by Cross-Alignment. on Advances in Neural Information Processing Systems (NeurIPS), 30, pp. 6833--6844 (2017).

\bibitem{Stamatatos2009}
Efstathios Stamatatos: A survey of modern authorship attribution methods. on Journal of the American Society for Information Science and Technology, 60(3), pp. 538--556 (2009).

\bibitem{Stamatatos2017}
Efstathios Stamatatos: Authorship Attribution Using Text Distortion. on Proceedings of the 15th Conference of the European Chapter of the Association for Computational Linguistics (EACL) (2017).

\bibitem{Yao2019}
Lili Yao, Nanyun Peng, Ralph Weischedel, Kevin Knight, Dongyan Zhao, and Rui Yan: Plan-and-Write: Towards Better Automatic Storytelling. on Proceedings of the AAAI Conference on Artificial Intelligence (AAAI), 33(01), pp. 7378--7385 (2019).

\bibitem{Yu2023}
Wenhao Yu, Dan Iter, Shuohang Wang, Yichong Xu, Mingxuan Ju, Soumya Sanyal, Chenguang Zhu, Michael Zeng, and Meng Jiang: Generate rather than retrieve: Large language models are strong context generators. on The Eleventh International Conference on Learning Representations (ICLR) (2023).
\end{thebibliography}
%

\end{document}